\documentclass[runningheads]{llncs}
\usepackage[T1]{fontenc}
\usepackage{graphicx}
\usepackage{hyperref}
\hypersetup{
    colorlinks=true,
    linkcolor=green,
    filecolor=magenta,
    urlcolor=cyan,
}
\usepackage{soul}
\soulregister\cite7
\usepackage[table,xcdraw]{xcolor}
\usepackage{multirow}
\usepackage{amsmath}
\usepackage{amssymb}
\usepackage{bm}
\usepackage{upgreek}
\usepackage{adjustbox}
\usepackage{makecell}
\usepackage{arydshln}

\usepackage{amsmath,amssymb,amsfonts}
\usepackage{graphicx}
\usepackage{textcomp}
\usepackage{multirow}
\usepackage{adjustbox}
\usepackage{makecell}
\usepackage{algpseudocode}

\usepackage{wrapfig}
\usepackage{marvosym}

\begin{document}
\title{Enhancing WSI-Based Survival Analysis with Report-Auxiliary Self-Distillation}
\titlerunning{Rasa}
%
\author{
Zheng Wang\inst{1} \and 
Hong Liu\inst{1} \and 
Zheng Wang\inst{1,2} \and 
Danyi Li\inst{3} \and 
Min Cen\inst{4} \and 
Baptiste Magnier \inst{5,6} \and 
Li Liang \inst{3} \and 
Liansheng Wang \inst{1}\textsuperscript{(\Letter)}
}
%
\authorrunning{Z. Wang et al.}
%

\institute{School of Informatics, Xiamen University, Xiamen, China\\
\email{\{zhengwang, liuhong, zwang\}@stu.xmu.edu.cn, lswang@xmu.edu.cn} \and
Shanghai Innovation Institution, Shanghai, China \and
Nanfang Hospital, Southern Medical University, Guangzhou, China\\
\email{lidanyi26@163.com,lli@smu.edu.cn}\and
School of Artificial Intelligence and Data Science, University of Science and Technology of China, Hefei, China\\
\email{cenmin0127@mail.ustc.edu.cn} \and
EuroMov Digital Health in Motion, Univ Montpellier, IMT Mines Ales, Ales, France\\
\email{baptiste.magnier@mines-ales.fr} \and
Service de Médecine Nucléaire, Centre Hospitalier Universitaire de Nîmes, Université de
Montpellier, Nîmes, France}


%
\maketitle              
\begin{abstract}
Survival analysis based on Whole Slide Images (WSIs) is crucial for evaluating cancer prognosis, as they offer detailed microscopic information essential for predicting patient outcomes.
However, traditional WSI-based survival analysis usually faces noisy features and limited data accessibility, hindering their ability to capture critical prognostic features effectively.
Although pathology reports provide rich patient-specific information that could assist analysis, their potential to enhance WSI-based survival analysis remains largely unexplored. 
To this end, this paper proposes a novel \textbf{R}eport-\textbf{a}uxiliary \textbf{s}elf-distill\textbf{a}tion (\textbf{Rasa}) framework for WSI-based survival analysis. 
First, advanced large language models (LLMs) are utilized to extract fine-grained, WSI-relevant textual descriptions from original noisy pathology reports via a carefully designed task prompt. 
Next, a self-distillation-based pipeline is designed to filter out irrelevant or redundant WSI features for the student model under the guidance of the teacher model's textual knowledge. 
Finally, a risk-aware mix-up strategy is incorporated during the training of the student model to enhance both the quantity and diversity of the training data.
Extensive experiments carried out on our collected data (CRC) and public data (TCGA-BRCA) demonstrate the superior effectiveness of Rasa against state-of-the-art methods. Our code is available at \href{https://github.com/zhengwang9/Rasa}{https://github.com/zhengwang9/Rasa}.


\end{abstract}
\section{Introduction}
%
%
Survival analysis based on Whole Slide Images (WSIs) is crucial for evaluating cancer prognosis, as they offer detailed microscopic information essential for predicting patient outcomes.
As the gold standard for cancer diagnosis and prognosis \cite{shao2021weakly}, WSIs capture critical features such as cellular structures, tumor microenvironment, and tissue phenotypes.
However, the effectiveness of traditional WSI-based survival analysis is often limited by two major challenges. 
First, the ultra-high resolution of WSIs introduces a vast number of irrelevant and redundant features, compromising analysis accuracy. 
Second, the acquisition of large-scale, high-quality data faces significant obstacles, including the need to meticulously label samples, privacy concerns, and extended follow-up periods. 

Previous studies have attempted to tackle the two challenges by employing data augmentation \cite{liu2024pseudo,wang2024rethinking,xiong2024takt} or de-noising directly on WSIs and labels \cite{chen2023rankmix,shao2021weakly}, while others have incorporated rich multimodal data to facilitate more sophisticated survival analysis \cite{chen2021multimodal,jaume2024modeling,xiong2024mome}. 
More recently, descriptions generated by advanced large language models (LLMs) have been introduced to enhance WSI-related tasks \cite{gou2025queryable,qu2024rise}. 
Compared with them, pathology reports offer richer patient-specific information, including high-level semantic descriptions of key findings, which could significantly aid analysis. 
However, the potential of leveraging these reports to enhance WSI-based survival analysis remains largely unexplored. This motivates our investigation into utilizing pathology reports to help overcome the two distinct limitations ({\it i.e.}, noisy features and limited data accessibility).

%
To this end, we proposes a novel \textbf{R}eport-\textbf{a}uxiliary \textbf{s}elf-distill\textbf{a}tion (\textbf{Rasa}) framework for WSI-based survival analysis. 
First, to tackle the issue of noise ({\it e.g.}, unmatched content with WSIs) in pathology reports, we employ advanced LLMs to transfer the noisy raw texts into detailed, WSI-aligned textual descriptions with a carefully crafted task-specific prompt.
%
Next, to facilitate the student model's focus on prognostically relevant information, we design a self-distillation pipeline that filters out irrelevant and redundant WSI features, guided by the teacher model's textual knowledge. 
%
Finally, we introduce a risk-aware mix-up strategy to enhance both data quantity and diversity during the student model training.
Our contributions are summarized as follows:
\begin{enumerate}
    \item We are the first to leverage pathology reports to improve WSI-based survival analysis, demonstrating the significant potential of integrating pathology reports with WSI analysis to advance computational pathology.
    \item We develop a novel \textbf{R}eport-\textbf{a}uxiliary \textbf{s}elf-distill\textbf{a}tion (\textbf{Rasa}) framework to enhance WSI-based survival analysis by addressing two core challenges, {\it i.e.}, noisy features and limited data accessibility in Sec. \ref{secMethod}. 
    \item We extensively evaluate our method on our collected data (CRC) and public data (TCGA-BRCA) in Sec. \ref{sec_exp}. The results demonstrate the superior performance of our method against the state-of-the-art (SOTA) methods.
\end{enumerate}

\begin{figure}
\centering
\includegraphics[width=\textwidth]{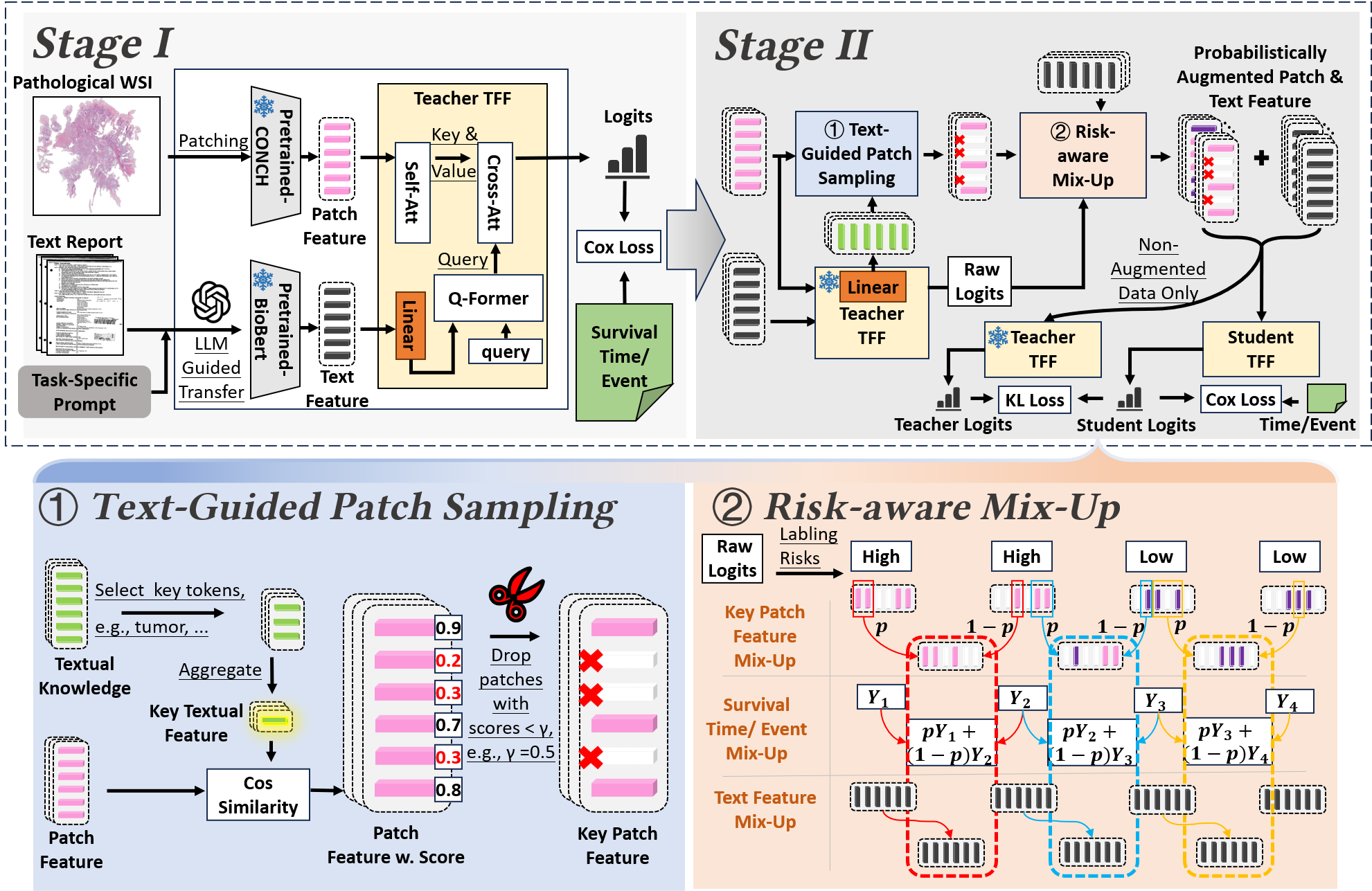}
\caption{The overview of Rasa framework.} \label{fig::method}
\end{figure}

\section{Method}\label{secMethod}

An overview of Rasa framework is available in the workflow presented in Fig. \ref{fig::method}.
It mainly includes two stages: Text-Fused Teacher Model Training and Tumor-Focused Student Model Training. In the first stage, we pre-process the raw data ({\it i.e.}, WSIs and pathology reports) into representative features and use them to train a teacher model. In the second stage, we employ the text-guided sampling module to filter noisy features and the risk-aware mix-up module to augment data. Subsequently, we train the student model under teacher model's guidance.

\subsection{Data Processing}
For each WSI $\mathcal{X}_i$, we use a pre-trained vision encoder of CONCH \cite{lu2024visual} to extract patch features $X_i=\{X_{ij}\}_{j=0}^{|X_i|}$.
For each pathology report $\mathcal{T}_i$, we first use GPT-4 \cite{achiam2023gpt} to transfer it into detailed, WSI-aligned textual descriptions and then encode them into token features $T_i=\{T_{ij}\}_{j=0}^{|T_i|}$ by a pre-trained BioClinicalBert \cite{wolf2020huggingfacestransformersstateoftheartnatural}. 
%
Specifically, we carefully designed a text prompt for GPT-4 \cite{achiam2023gpt} to emphasize the details of microscopic visual characteristics in WSIs and eliminate WSI-irrelevant information ({\it e.g.}, lymph node information, immunohistochemistry results, and certain genetic data). This facilitates the interaction between textual information and WSIs, as empirically evidenced by Table \ref{tab2} and Fig. \ref{Fig:sim}. 
%
%
%

\subsection{Text-Fused Teacher Model Training}
After processing data, we use all extracted features ({\it i.e.}, patch feature $X_i$ and text feature $T_i$) to pre-train a Text-Fused Former (TFF) model as the teacher ({\it i.e.}, Stage I in Fig. \ref{fig::method}). Concretely, $T_i$ is converted into the teacher's textual knowledge $T_{i}^{(proj)}$ via a projector ({\it i.e.}, the orange Linear in Fig. \ref{fig::method}) and refined with a Q-Former \cite{li2023blip}. For $X_i$, we directly encode it via a self-attention module. By using $T_i'$ as the query and $X_i'$ as the key and value, we compute cross-attention between $T_i'$ and $X_i'$ to summarize the task-essential information, which is pooled and passed through the head for predictions. The training objective is presented in Eq. (\ref{Equ:Cox_t}), and the forward process of the TFF model is outlined as follows:
\begin{align}
    T_i'=\text{Q-Former}(T_{i}^{(proj)}), \quad T_{i}^{(proj)}&=\text{Linear}(T_i), \quad X_i'=\text{self-attn}(X_i),\notag\\
    y_{i}=\text{head}(\text{Pool}(Z_i)&), \quad Z_i=\text{cross-attn}(T_i', X_i', X_i'). \label{eq_forward}
\end{align}
%
%
%
\subsection{Tumor-Focused Student Model Training}\label{sec_student}
\subsubsection{Text-Guided Patch Sampling:} \label{Sec:PatchSp}
While the extremely high resolution of WSIs benefits survival analysis by offering rich information, it also introduces numerous irrelevant and redundant patches that can distract the model and compromise its performance.
Since pathologists typically draw conclusions based on lesion regions ({\it i.e.}, cancerous areas) in clinical practice \cite{connolly2003role}, we propose to filter out noise in WSIs by leveraging the teacher model's textual knowledge ({\it e.g.}, Block \textcircled{1} in Fig. \ref{fig::method}). 
First, we select token features $\mathcal{S}_i=\{T_{ij}^{(proj)}| j \text{ is selected}\}$ from the teacher's textual knowledge $T_{i}^{(proj)}$ corresponding to manually specified keywords ({\it e.g.}, ``tumor'' or ``cancer``) in tokenized input texts. 
Next, we average selected features into the key textual feature \(T_{i}^{(key)} = \frac{1}{|\mathcal{S}_i|}\sum_{j\in \mathcal{S}_i}T_{ij}^{(proj)}\). Finally, we filter out patches dissimilar to the key textual feature using cosine similarity:
\begin{align}
    X_i^{(key)} = \left\{X_{ij} \mid \frac{X_{ij} \cdot T_{i}^{(key)}}{\|X_{ij}\|_2\|T_{i}^{(key)}\|_2} \ge \gamma, \quad X_{ij} \in X_i\right\},
\end{align}
where \(\gamma\) is a pre-defined threshold. This strategy effectively filters out a large number of noisy patches by retaining only key patches strongly associated with tumor regions, as shown in Fig. \ref{Fig:sim}, and it also enhances efficiency by eliminating time-consuming, expertise-dependent manual labeling of cancerous areas.
\subsubsection{Risk-aware Mix-up:} 
To tackle the challenge of limited data accessibility, we consider employing mix-up-based data augmentation to improve the quantity and diversity of the training data. However, directly mixing up pairs of samples ({\it e.g.}, WSIs, labels and reports) without recognizing their risks could potentially yield misleading fusion. For example, it would be improper to associate a mixed WSI that contains high-risk patches with a low-risk text report. 
To this end, we first use the pre-trained teacher model to label the samples' risk $ r_{i}=\mathbb{I}(s_i\ge s_{\text{medium}})$, where $s_{i}=\text{sigmoid}(y_{i})$ and $s_{\text{medium}}$ is the medium number of $\{s_i\}_{i=1}^N$ over the training set.
Then, we respectively mix each sample's text features $T_a$, key patch features $X_a^{(key)}$, and labels $Y_a$ with another one's ({\it e.g.}, $(T_b, X_b^{(key)}, Y_b)$) into a new sample $(T_{\text{mix}}, X_{\text{mix}}, Y_{\text{mix}})$ with a probability $p_{\text{aug}}$. 
For the text feature, we directly set $T_{\text{mix}}=T_a$ for pairs of samples with homogeneous risks ({\it e.g.}, high-high and low-low risks) while using $T_{\text{mix}}=T_{\text{argmax}(r_a, r_b)}$ for pairs of samples with heterogeneous risks ({\it e.g.}, high-low risks). 
This is because high-risk samples often provide more critical characteristics than low-risk ones in survival analysis. For patch features, we randomly select $100*p_{\text{mix}}\%$ from $X_a^{(key)}$ and $100*(1-p_{\text{mix}})\%$ from $X_b^{(key)}$ and combine them into $X_{mix}$. 
Compared with previous bag mix-up strategies \cite{chen2023rankmix,liu2024pseudo}, we only mix the key patches selected by the textual knowledge, as mixing low-information patches causes low efficiency in increasing data diversity. 
For the labels $Y=(c,t)$ --{\it e.g.}, the censoring status $c$ and survival time $t$--  we adopt a soft-mixing strategy to ensure that the mixed label accurately reflects the contribution of both participants, as is computed below: 
\begin{equation}
\begin{aligned}
    c_{mix} = (1 - p_{mix})c_a + p_{mix}c_b , \quad
    t_{mix} = (1 - p_{mix})t_a + p_{mix}t_b.
\end{aligned}
\end{equation}

\subsection{Training Procedure} 
The task objective is to minimize the Cox loss \cite{yao2020whole} as defined below:
\begin{equation}\label{Equ:Cox_t}
    \mathcal{L}_{cox} = -\sum_{i=1}^N \delta_i\left(y_i - \log\sum_{j\in R(t_i)} e^{y_j}\right),
\end{equation}
where $y_i$ represents the output of the model, and $R(t_i)$ is the risk set at time $t_i$.
Besides, we introduce Kullback-Leibler (KL) divergence \cite{van2014renyi} to leverage the teacher model to guide the student model on non-augmented samples as:
\begin{equation}
    \mathcal{L}_{KL} = D_{KL}(y_{i, student} \| y_{i, teacher}),
\end{equation}
as the teacher model might yield unreliable results on unseen augmented samples. The student's objective $\mathcal{L}_{cox}+\lambda \mathcal{L}_{KL}^{(non-aug)}$ enables it to additionally learn from the teacher's refined knowledge, where $\lambda$ is set to balance the two objectives. 



\section{Experiment}\label{sec_exp}
\subsection{Experimental Settings}
\subsubsection{Datasets:}
The experiments were conducted on a Colorectal Cancer (CRC) cohort comprising 302 cases collected from a collaborating hospital, and a publicly available Breast Invasive Carcinoma cohort from The Cancer Genome Atlas (TCGA-BRCA) \cite{hutter2018cancer}, consisting of 331 cases. We employed a 5-fold cross-validation where the train/validation/test ratio is 0.6/0.2/0.2 within each trial.
%
%
%
%
\subsubsection{Metric:}
We use the Concordance Index (CI) \cite{wang2019machine} as the metric to measure the performance in predicting survival outcomes. We fairly report the averaged testing results of the models that optimally perform on the validation set.
%
%
%
%
%
\subsubsection{Implementation:}
We use Adam optimizer to train the model  \cite{kingma2014adam} for 60 epochs with a fixed batch size of 8 and a learning rate of \(1 \times 10^{-5}\). $\lambda$ is optimally tuned to \(1 \times 10^{-2}\) and \(1 \times 10^{-5}\), respectively for CRC and TCGA-BRCA datasets. 
%
\begin{table}[t]
\caption{The performance of our model compared with SOTA methods.}\label{tab1}
\centering
\begin{adjustbox}{width=.82\textwidth}
\begin{tabular}{c|c|c|c}
\hline
Type & Method & CRC & TCGA-BRCA \\
\hline
Vision-only & ABMIL \cite{ilse2018attention} & $0.5132_{\pm 0.0982}$ & $0.6368_{\pm 0.0437}$ \\
 & PatchGCN \cite{chen2021whole} & $0.5474_{\pm 0.1144}$ & $0.6372_{\pm 0.0611}$ \\
 & TransMIL \cite{shao2021transmil} & $0.5348_{\pm 0.0787}$ & $0.5934_{\pm 0.0232}$ \\
 & DSMIL \cite{li2021dual} & $0.5234_{\pm 0.1256}$ & $0.6284_{\pm 0.0509}$ \\
 & MambaMIL \cite{yang2024mambamil} & $0.5416_{\pm 0.0954}$ & $0.6366_{\pm 0.0149}$ \\
\hline
Vision \& Text & QPMIL-VL \cite{gou2025queryable} & $0.5748_{\pm 0.0733}$ & $0.5826_{\pm 0.0517}$ \\
 & TOP \cite{qu2024rise} & $0.5488_{\pm 0.0647}$ & $0.5434_{\pm 0.0291}$ \\
 & MCAT \cite{han2024mscpt} & $0.5592_{\pm 0.1020}$ & $0.6198_{\pm 0.0520}$ \\
\hline
Bag Mix-up & PseMix \cite{liu2024pseudo} & $0.5824_{\pm 0.1018}$ & $0.6500_{\pm 0.0432}$ \\
 & RankMix \cite{chen2023rankmix} & $0.5262_{\pm 0.1349}$  & $0.6120_{\pm 0.0803}$ \\
\hline
 & Rasa (Ours) & $\textbf{0.6834}_{\pm 0.1331}$ & $\textbf{0.6972}_{\pm 0.0500}$ \\
 \hline
\end{tabular}
\end{adjustbox}
\end{table}

\subsection{Comparison with SOTA Methods}
%
%
%
We compare our methods with three types of baselines: 
{\it i)} vision-only models \cite{ilse2018attention,chen2021whole,shao2021transmil,li2021dual,yang2024mambamil}, 
{\it ii)} vision-language (VL) models \cite{gou2025queryable,qu2024rise,chen2021multimodal}, and 
{\it iii)} bag mix-up strategies \cite{chen2023rankmix,liu2024pseudo}. The results in Table \ref{tab1} suggest the superiority of our approach on both datasets in survival prediction tasks.
First, vision-only models achieve moderate performance, revealing the limitations of relying solely on noisy visual information for capturing nuanced details essential for accurate survival prediction. 
%
Second, vision-language (VL) models show only marginal improvements on CRC and a slight decline on TCGA-BRCA against vision-only models. We attribute this to the simplistic textual information ({\it e.g.}, generic class names and GPT-generated descriptions) used by these VL models, lacking sufficient slide-specific details for further improvement.
Third, among the bag mix-up augmentation strategies,  PseMix (Pseudo-bag mixup augmentation \cite{liu2024pseudo}) stands out as the best-performing baseline, demonstrating the effectiveness of mix-up techniques in improving model performance.  The poor performance of RankMix \cite{chen2023rankmix} may be due to its heavy reliance on the ability of the teacher model.
Finally, our Rasa achieves the highest performance on both datasets, highlighting the effectiveness of generating slide-specific textual descriptions and integrating them into the decision process.


\subsection{Ablation Studies}
\subsubsection{Impact of Text:}
To validate our designed textual modality, we tested various alternatives in Table \ref{tab2}: no text (w/o Text), ``Tumor cells'',  GPT-generated descriptions (GPT Text), original pathology reports (Original Report), and CONCH text embeddings (CONCH Text-Encoder). 
Removing text ({\it i.e.}, w/o text) yields the worst results while using simple ``Tumor cells'' shows non-trivial improvement, highlighting the indispensable importance of texts. 
GPT Text achieves suboptimal performance by offering rich textual context. Although Original Report introduces more patient-specific information than GPT Text while similarly offering context, its effectiveness is limited by noise. CONCH text embeddings also underperform our method that uses BioClinicalBert. 
Our approach, leveraging LLMs for precise, contextually aligned text descriptions, achieves the best performance, demonstrating the effectiveness of advanced text processing in pathological image analysis. These findings underscore the importance of sophisticated text integration for accurate and robust survival prediction.
\begin{table}[!t]
  \centering

  \begin{minipage}{0.445\textwidth}

    \caption{Results of textual variates}\label{tab2}
    \centering
    \begin{adjustbox}{width=\textwidth}
    \begin{tabular}{c|c|c}
    \hline
    Config & CRC & TCGA-BRCA \\
    \hline
     w/o Text & $0.5566_{\pm 0.1137}$ & $0.6362_{\pm 0.0335}$  \\
     ’Tumor cells' & $0.6020_{\pm 0.0988 }$ & $0.6618_{\pm 0.0735}$  \\
     GPT Text & $0.6450_{\pm 0.0823}$ & $0.6636_{\pm 0.0407}$ \\
     Original Report & $0.6394_{\pm 0.0973}$ & $0.6564_{\pm 0.0224}$ \\
     CONCH Text-Encoder & $0.6178_{\pm 0.1529}$ & $0.6354_{\pm 0.0532}$ \\
    \hline
     Ours & $\textbf{0.6834}_{\pm 0.1331}$ & $\textbf{0.6972}_{\pm 0.0500}$ \\
    \hline
    \end{tabular}
    \end{adjustbox}
    \end{minipage}
 \hspace{0.01\textwidth}
 \begin{minipage}{0.525\textwidth}

    \caption{Ablation study on sub-modules}\label{tab::ablation}
    \centering
    \begin{adjustbox}{width=\textwidth}
    \begin{tabular}{c|c|c}
    \hline
    Config & CRC & TCGA-BRCA \\
    \hline
    Teacher Model & $0.6196_{\pm 0.1336}$ & $0.6494_{\pm 0.0269}$ \\
    w/o Loss KL \& Mix-up & $0.6678_{\pm 0.1663}$ & $0.6618_{\pm 0.0620}$ \\
    w/o Mix-up & $0.6746_{\pm 0.1465}$ & $0.6750_{\pm 0.0510}$ \\ 
     w/o Loss KL & $0.6726_{\pm 0.1316}$ & $0.6908_{\pm 0.0430}$ \\
    \hline
     Ours & $\textbf{0.6834}_{\pm 0.1331}$ & $\textbf{0.6972}_{\pm 0.0500}$ \\
    \hline
    \end{tabular}
    \end{adjustbox}
\end{minipage}
\end{table}

\subsubsection{Effect of Sub-Module:}
We evaluate the impact of each sub-module in Rasa through ablation studies in Table \ref{tab::ablation}. The results show that directly using the teacher model yields the poorest performance. Removing either the Mix-up module or the teacher guidance from $\mathcal{L}_{KL}$ leads to a performance drop, while removing both results in a more significant decline. This indicates the effectiveness of each sub-module and the importance of their collaborative interaction.

\begin{figure}[b!]
\centering
\includegraphics[width=.9\textwidth]{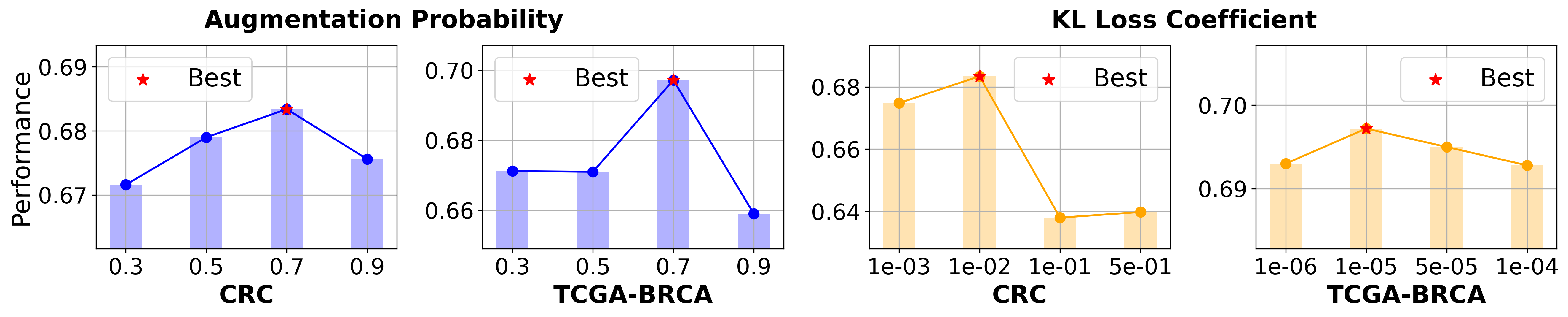} \vspace{-.2cm}
\caption{Impact of the augmentation probability $p_{aug}$ and the coefficient $\lambda$} \label{Tabel::hyper-params}
\end{figure}

\subsubsection{Effect of Hyper-Parameters:}
We investigate the impact of $p_{aug}$ ({\it i.e.}, the augmentation probability), $\lambda$ ({\it i.e.}, the coefficient of $\mathcal{L}_{KL}$) in Fig. \ref{Tabel::hyper-params}. 
For $p_{aug}$, a lower value may lead to insufficient diversity of data, while a higher one could introduce excessive noise, potentially degrading the model's performance. The optimal value of $p_{aug}$ is $0.7$ for both the CRC and TCGA-BRCA datasets.
For $\lambda$, optimal performance was achieved with values of \(1 \times 10^{-2}\) and \(1 \times 10^{-5}\) for the CRC and TCGA-BRCA datasets, respectively. 
%
\subsection{Visualization}
\begin{figure}[t]
\centering
\includegraphics[width=.83\textwidth]{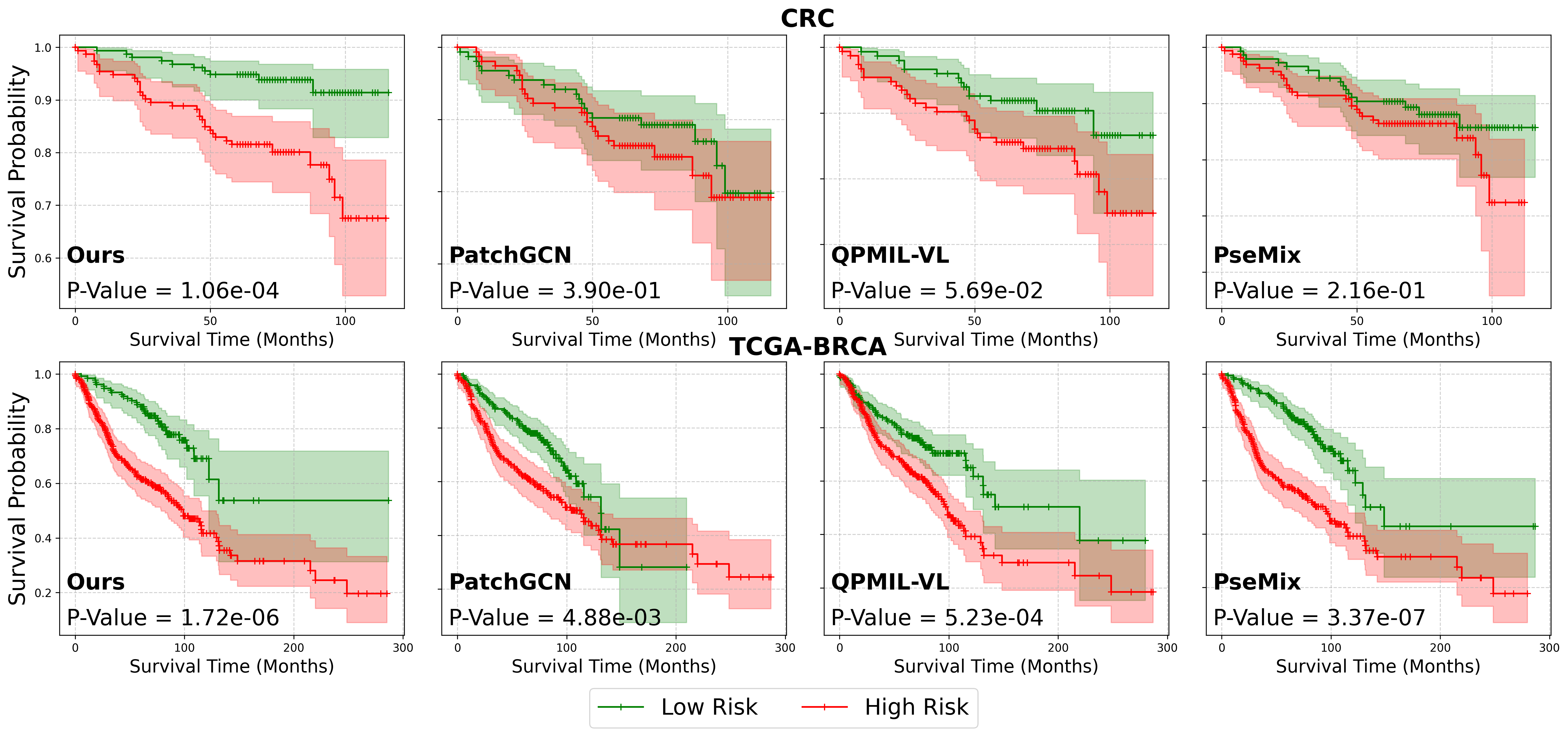} \vspace{-.2cm}
\caption{Kaplan-Meier curves for predicted high-risk (red) and low-risk (green) groups on the test sets of the CRC (top) and TCGA-BRCA (bottom) datasets
} \label{Fig:KM}
\end{figure}
\subsubsection{Kaplan–Meier Analysis:}
We further validate the effectiveness of our method via Kaplan-Meier (KM) curves in Fig. \ref{Fig:KM}. We follow \cite{cheng2018identification} to divide patients in the test set into low-risk and high-risk groups based on the median risk score from the training set. 
The statistical significance of the survival time differences between these groups was evaluated using the log-rank test, with a p-value below 0.01 indicating statistical significance.
Compared with other SOTA methods, our method achieved remarkably low p-values on both datasets while demonstrating a clear and robust separation between low-risk and high-risk groups. 
%
%

\begin{figure}[b!]
\centering
\includegraphics[width=.83\textwidth]{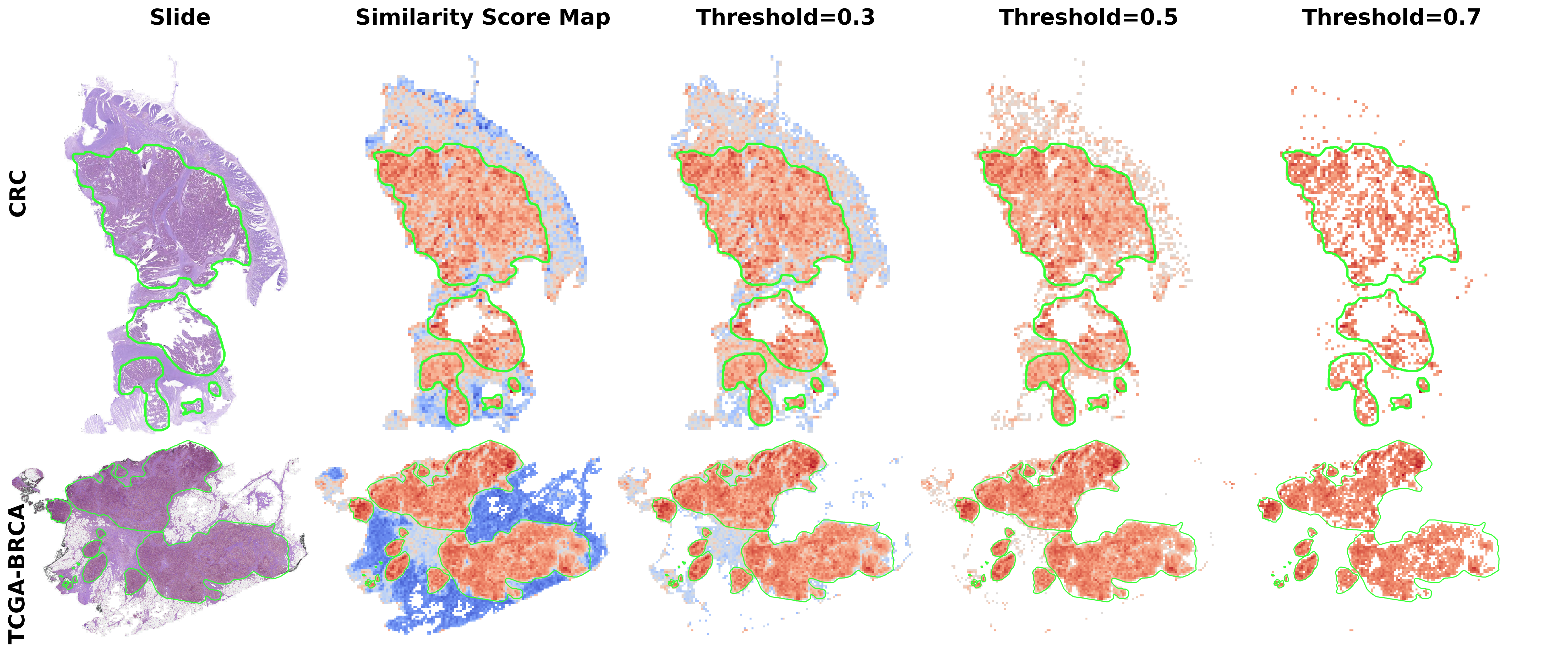} \vspace{-.2cm}
\caption{Text-patch similarity score maps and preserved patches with different $\gamma$.} \label{Fig:sim}
\end{figure}
\subsubsection{Effect of Threshold:}
%
%
%
%
%
%

The text-patch similarity score map with different values of $\gamma$ (the patch sampling filter threshold) is shown in Fig. \ref{Fig:sim}.
%
Regions with high text-patch similarity predominantly overlap with cancerous areas (\textit{i.e.}, green-lined regions), demonstrating that the teacher model's refined text features effectively align the embeddings of text keywords with patch embeddings. For another thing, the impact of varying thresholds on patch filtering is evident, as increasing the threshold retains more cancerous patches but introduces noisy patches. We selected a threshold of 0.5 in experiments, as it optimally balances the retention of cancerous regions with minimal inclusion of non-cancerous patches.
%

\section{Conclusion}
In this paper, we propose a \textbf{R}eport-\textbf{a}uxiliary \textbf{s}elf-distill\textbf{a}tion (\textbf{Rasa}) framework to address two core challenges in WSI-based survival analysis: noisy features and limited data accessibility. By leveraging advanced LLMs and carefully designing modules, we successfully enhanced WSI-based survival analysis with the assistance of reports. Extensive experiments on CRC and TCGA-BRCA datasets confirm the superiority of Rasa. We plan to fully release the power of the report-auxiliary data enhancement technique in more WSI analysis tasks in the future. 
\begin{credits}
\subsubsection{\ackname} This work was supported by National Natural Science Foundation of China (Grant No. 62371409) and Fujian Provincial Natural Science Foundation of China (Grant No. 2023J01005).
\subsubsection{\discintname}
The authors have no competing interests to declare that are relevant to the content of this article.
\end{credits}
%
%
\bibliographystyle{splncs04}
\bibliography{Paper-1264}

\end{document}